\documentclass{article} 
\usepackage{iclr2018_conference,times}

\usepackage{url}
\usepackage{amsfonts,epsf,psfrag,amssymb,amsthm,bbm}
\graphicspath{ {Plots/} }
\usepackage[dvips]{graphicx}
\usepackage{hyperref}
\usepackage{graphicx} 
\usepackage{subcaption}
\usepackage{xspace} 
\usepackage{natbib}
\usepackage{algorithm}
\usepackage{algorithmicx}
\usepackage{algpseudocode}
\usepackage{hyperref}
\usepackage[disable, colorinlistoftodos, textwidth=20mm]{todonotes}
\definecolor{citrine}{rgb}{0.89, 0.82, 0.04}
\definecolor{blued}{RGB}{70,197,221}

\usepackage[utf8]{inputenc} 
\usepackage[T1]{fontenc}    
\usepackage{hyperref}       
\usepackage{url}            
\usepackage{booktabs}       
\usepackage{nicefrac}       
\usepackage{microtype}      
\usepackage{amsmath,epsf,psfrag,amssymb}
\usepackage{epsf,psfrag,amsfonts, xcolor,array}
\usepackage{amsthm}
\usepackage{multirow}
\usepackage{cases}
\usepackage{verbatim}
\usepackage{xspace}
\usepackage{bbm}
\usepackage{mathrsfs}
\usepackage{setspace}
\usepackage[fleqn,tbtags]{mathtools}
\usepackage{tikz}
\usepackage{todonotes}
\usepackage{bm}

\newcommand{\sap}{\textsc{\small{SAP}}\xspace}

\title{Stochastic activation pruning for\\ robust adversarial defense}

\author{
Guneet S. Dhillon$^{1,2}$,
Kamyar Azizzadenesheli$^{3}$,
Zachary C. Lipton$^{1,4}$,
\\\textbf{
Jeremy Bernstein$^{1,5}$,
Jean Kossaifi$^{1,6}$,
Aran Khanna$^1$,
Anima Anandkumar$^{1,5}$
}\\
$^1$Amazon AI,
$^2$UT Austin,
$^3$UC Irvine,
$^4$CMU,
$^5$Caltech,
$^6$Imperial College London
\\ \hfill\texttt{
guneetdhillon@utexas.edu,
kazizzad@uci.edu,
zlipton@cmu.edu,
}\\ \hfill\texttt{
bernstein@caltech.edu, jean.kossaifi@imperial.ac.uk,
}\\ \hfill\texttt{
aran@arankhanna.com,
anima@amazon.com
}
}

\iclrfinalcopy 
\begin{document}

\maketitle

\begin{abstract}
Neural networks are known 
to be vulnerable to adversarial examples. Carefully chosen perturbations to real images,
while imperceptible to humans, 
induce misclassification and
threaten the reliability of deep learning systems in the wild. 
To guard against adversarial examples, 
we take inspiration from game theory
and cast the problem as a minimax zero-sum game 
between the adversary and the model.
In general, for such games, the optimal strategy  for both players requires a stochastic policy, also known as a \emph{mixed strategy}.
In this light, we propose \emph{Stochastic Activation Pruning} (\sap), 
a mixed strategy for adversarial defense. 
\sap prunes a random subset of activations 
(preferentially pruning those 
with smaller magnitude) 
and scales up the survivors to compensate.
We can apply \sap to pretrained networks, 
including adversarially trained models, 
without fine-tuning,
providing robustness against adversarial examples. 
Experiments demonstrate that
\sap confers robustness against attacks, 
increasing accuracy and preserving calibration.
\end{abstract}

\section{Introduction}
\label{sec:introduction}
While deep neural networks have emerged as 
dominant tools for supervised learning problems,
they remain vulnerable to adversarial examples \citep{szegedy2013intriguing}.
Small, carefully chosen perturbations to input data 
can induce misclassification with high probability.
In the image domain,
even perturbations so small as to be imperceptible to humans
can fool powerful convolutional neural networks \citep{szegedy2013intriguing, goodfellow2014explaining}.
This fragility presents an obstacle
to using machine learning in the wild.
For example, a vision system vulnerable to adversarial examples 
might be fundamentally unsuitable
for a computer security application.
Even if a vision system 
is not explicitly used for security, 
these weaknesses might be critical.
Moreover, these problems seem unnecessary.
If these perturbations are not perceptible
to people, why should they fool a machine?

Since this problem was first identified,
a rapid succession of papers have 
proposed various techniques
both for \emph{generating} 
and for \emph{guarding against} adversarial attacks.
\citet{goodfellow2014explaining} introduced
a simple method for quickly producing adversarial examples 
called \emph{the fast gradient sign method} (FGSM). 
To produce an adversarial example using FGSM,
we update the inputs by taking one step 
in the direction of the \emph{sign} of the gradient of the loss with respect to the input. 

To defend against adversarial examples
some papers propose training the neural network
on adversarial examples themselves,
either using the same model \citep{goodfellow2014explaining, madry2017towards},
or using an ensemble of models \citep{tramer2017ensemble}.
Taking a different approach, \citet{nayebi2017biologically}
draws inspiration from biological systems.
They propose that to harden neural networks
against adversarial examples,
one should learn flat, compressed representations
that are sensitive to a minimal number of input dimensions.

This paper introduces \emph{Stochastic Activation Pruning} (\sap), a method
for guarding pretrained networks
against adversarial examples.
During the forward pass,
we stochastically prune a subset of the activations in each layer,
preferentially retaining activations with larger magnitudes.
Following the pruning, we scale up the surviving activations 
to normalize the dynamic range 
of the inputs to the subsequent layer.
Unlike other adversarial defense methods,
our method can be applied post-hoc
to pretrained networks
and requires no additional fine-tuning.

\section{Preliminaries}
\label{sec:preliminaries}
We denote an $n$-layered neural network 
$h: \mathcal{X}\rightarrow Y$ 
as a chain of functions 
$h=h^{n}\circ h^{n-1}\circ\ldots \circ h^{1}$, 
where each $h^i$ consists of a linear transformation $W^i$ followed by a non-linearity $\phi^i$.
Given a set of nonlinearities and weight matrices, 
a neural network provides a nonlinear mapping
from inputs $x\in\mathcal{X}$
to outputs $\hat{y}\in \mathcal{Y}$, 
i.e.
\begin{align*}
\hat{y} := h(x) = \phi^n(W^n\phi^{n-1}(W^{n-1}\phi^{n-2}(\ldots\phi^1(W^1x)))).
\end{align*}

In supervised classification and regression problems, 
we are given a data set $\mathcal{D}$ of pairs $(x,y)$,
where each pair is drawn from an unknown joint distribution. 
For the classification problems, 
$y$ is a categorical random variable,
and for regression, $y$ is a real-valued vector. 
Conditioned on a given dataset, network architecture,
and a loss function such as cross entropy, 
a learning algorithm, e.g. stochastic gradient descent,
learns parameters 
$\theta:=\lbrace W^i\rbrace_{i=1}^{n}$ in order to minimize the loss. 
We denote $J(\theta,x,y)$ as the loss of a learned network, 
parameterized by $\theta$, 
on a pair of $(x,y)$. 
To simplify notation, we focus on classification problems, although our methods are broadly applicable.

Consider an input $x$ that is correctly classified by the model $h$. 
An adversary seeks to apply a small additive perturbation, $\Delta x$, 
such that
$h(x)\neq h(x+\Delta x)$,
subject to the constraint that the perturbation 
is imperceptible to a human.
For perturbations applied to images,
the $l_{\infty}$-norm is considered a better measure 
of human perceptibility 
than the more familiar $l_{2}$ norm \citet{goodfellow2014explaining}.
Throughout this paper,
we assume that
the manipulative power of the adversary, 
the perturbation $\Delta x$, 
is of bounded norm
$\|\Delta x\|_\infty\leq \lambda$.
Given a classifier, one common way to generate an adversarial example 
is to perturb the input in the direction that increases the cross-entropy loss.
This is equivalent to minimizing the probability assigned to the true label.
Given the neural network $h$, 
network parameters $\theta$, 
input data $x$, and corresponding true output $y$,
an adversary could create a perturbation $\Delta x$ as follows
\begin{align}\label{eq:opt}
\Delta x=\arg\max_{\|r\|_\infty\leq\lambda}J(\theta,x+r,y),
\end{align}
Due to nonlinearities in the underlying neural network, and therefore of the objective function $J$, the optimization Eq.~\ref{eq:opt}, in general, can be a non-convex problem. 
Following \citet{madry2017towards,goodfellow2014explaining}, we use the first order approximation of the loss function 
\begin{align*}
\Delta x=\arg\max_{\|r\|_\infty\leq\lambda}[J(\theta,x,y)+r^{\top}\mathcal{J}(\theta,x,y)],&&\text{where }\mathcal{J}=\nabla_x J.
\end{align*}
The first term in the optimization is not a function of the adversary perturbation, therefore reduces to
\begin{align*}
\Delta x=\arg\max_{\|r\|_\infty\leq\lambda}r^{\top}\mathcal{J}(\theta,x,y).
\end{align*}
An adversary
chooses $r$ to be in the direction of sign of $\mathcal{J}(\theta,x,y)$, i.e. $\Delta x=\lambda \cdot \text{sign}(\mathcal{J}(\theta,x,y))$.
This is the
FGSM technique due to \citet{goodfellow2014explaining}.
Note that FGSM requires 
an adversary to access the model
in order to compute the gradient.

\section{Stochastic activation pruning}
\label{sec:sap}
Consider the defense problem
from a game-theoretic perspective \citep{osborne1994course}. 
The adversary designs a policy in order to maximize the defender's loss,
while knowing the defenders policy. 
At the same time defender aims to come up with a strategy to minimize the maximized loss. 
%
%
Therefore, we can rewrite Eq.~\ref{eq:opt} as follows
\begin{align}\label{eq:minimax}
\pi^*~,~\rho^* := \arg\min_{\pi}\max_{\rho}\mathbb{E}_{p\sim\pi,r\sim\rho}\left[ J(M_{p}(\theta),x+r,y)\right],
\end{align}
where $\rho$ is the adversary's policy, which provides $r\sim \rho$ in the space of bounded (allowed) perturbations (for any $r$ in range of $\rho$, $\|r\|_\infty\leq\lambda$) and $\pi$ is the defenders policy which provides $p\sim\pi$, an instantiation of its policy.
The adversary's goal is to maximize the loss of the defender 
by perturbing the input under a strategy $\rho$ 
and the defender's goal is to minimize the loss 
by changing model parameters $\theta$ to
$M_{p}(\theta)$ under strategy $\pi$.
The optimization problem in Eq. \ref{eq:minimax} 
is a \textit{minimax} zero-sum game 
between the adversary and defender 
where the optimal strategies $(\pi^*,\rho^*)$, 
in general, 
are mixed Nash equilibrium, 
i.e. stochastic policies. 

Intuitively, the idea of \sap is 
to stochastically drop out nodes in each layer 
during forward propagation.
We retain nodes with probabilities proportional 
to the magnitude of their activation 
and scale up the surviving nodes 
to preserve the dynamic range 
of the activations in each layer. 
Empirically, the approach preserves 
the accuracy of the original model.
Notably, the method can be applied post-hoc 
to already-trained models.

Formally, assume a given pretrained model, 
with activation layers (\textit{ReLU, Sigmoid, etc.}) and input pair of $(x,y)$.
For each of those layers, 
\sap converts the activation map
to a multinomial distribution,
choosing each activation
with a probability proportional to its absolute value.
In other words,
we obtain the multinomial distribution
of each activation layer 
with $L_1$ normalization of their absolute values onto a $L_1$-ball simplex.
Given the $i$'th layer activation map, 
$h^{i}\in\mathbb{R}^{a^{i}}$, 
the probability of sampling the $j$'th
activation with value $(h^{i})_{j}$
is given by
\begin{align*}
p^{i}_{j}=\frac{|(h^{i})_{j}|}{\sum_{k=1}^{a^{i}}{|(h^{i})_{k}|}}.
\end{align*} 
We draw random samples with replacement 
from the activation map 
given the probability distribution described above.
This makes it convenient to determine 
whether an activation would be sampled at all.
If an activation is sampled, 
we scale it up by the inverse of the probability 
of sampling it over all the draws. 
If not, we set the activation to $0$. 
In this way, \sap preserves inverse propensity scoring of each activation.
Under an instance $p$ of policy $\pi$, 
we draw $r^{i}_{p}$ samples with replacement 
from this multinomial distribution.
The new activation map, $M_{p}(h^{i})$ 
is given by
\begin{align*}
M_{p}(h^{i})=h^{i}\odot m^{i}_{p},&&
(m^{i}_{p})_{j}=\frac{\mathbb{I}((h^{i})_{j})}{1-(1-p^{i}_{j})^{r^{i}_{p}}},
\end{align*}
where $\mathbb{I}((h^{i})_{j})$ 
is the indicator function
that returns $1$ if $(h^{i})_{j}$ was sampled 
at least once,
and $0$ otherwise.
The algorithm is described in Algorithm \ref{alg:sap}.
In this way, the model parameters
are changed from $\theta$ to $M_{p}(\theta)$,
for instance $p$ under policy $\pi$, 
while the reweighting $1-(1-p^{i}_{j})^{r^{i}_{p}}$ preserves $\mathbb{E}_{p\sim\pi}[M_{p}(h^{i})_{j}] =(h^{i})_{j}$.
If the model was linear,
the proposed pruning method
would behave the same way as the original model
in expectation.
In practice, we find that even with the non-linearities in deep neural networks, for sufficiently many examples, 
\sap performs similarly to the un-pruned model.
This guides our decision to apply \sap to pretrained models without performing fine-tuning.

\algtext*{EndIf}
\algtext*{EndFor}
\begin{algorithm}[t]
\caption{Stochastic Activation Pruning (\sap)}
\begin{algorithmic}[1]
\State \textbf{Input:} input datum $x$, neural network with $n$ layers, with $i^{th}$ layer having weight matrix $W^i$, non-linearity $\phi^i$ and number of samples to be drawn $r^i$.
\State $h^{0}\leftarrow x$
\For{\textbf{each} layer $i$}
	\State $h^{i}\gets\phi^{i}(W^{i}h^{i-1})$\Comment{activation vector for layer $i$ with dimension $a^{i}$}
    \State $p^{i}_{j}\leftarrow\frac{|(h^{i})_{j}|}{\sum_{k=1}^{a^{i}}{|(h^{i})_{k}|}},~\forall j \in \{1,\ldots,a^{i}\}$ \Comment{activations normalized on to the simplex} 
    \State $S\leftarrow\{\}$ \Comment{set of indices not to be pruned}
    \State \textbf{repeat} $r^{i}$ \textbf{times}\Comment{the activations have $r^i$ chances of being kept}
    \State $\quad\,\,\,$Draw $s \sim \mathrm{categorical}(p^{i})$ \Comment{draw an index to be kept}
    \State $\quad\,\,$ $S\leftarrow S\cup \{s\}$ \Comment{add index $s$ to the keep set}
     \For{\textbf{each} $j \notin S$}
     \State $(h^{i})_{j} \gets 0$ \Comment{prune the activations not in $S$}     
    \EndFor
    \For{\textbf{each} $j \in S$}
        	\State $(h^{i})_{j} \gets \frac{(h^{i})_{j}}{1-(1-p^{i}_{j})^{r^{i}}}$ \Comment{scale up the activations in $S$}
    \EndFor
\EndFor
\State \Return $h^{n}$
\end{algorithmic}
\label{alg:sap}
\end{algorithm}

\subsection{Advantage against adversarial attack}

We attempt to explain 
the advantages of \sap
under the assumption 
that we are applying it to a pre-trained model that achieves high generalization accuracy.
For instance $p$ under policy $\pi$, 
if the number of samples drawn for each layer $i$, $r^{i}_{p}$, is large, 
then fewer parameters of the neural network are pruned, 
and the scaling factor gets closer to $1$. 
Under this scenario, the stochastically pruned model 
performs almost identically to the original model. 
The stochasticity is not advantageous in this case, 
but there is no loss in accuracy in the pruned model 
as compared to the original model.

On the other hand, with fewer samples in each layer, $r^{i}_{p}$, 
a large number of parameters of the neural network are pruned. 
Under this scenario, 
the \sap model's accuracy will drop 
compared to the original model's accuracy.
But this model is stochastic and has more freedom to deceive the adversary.
So the advantage of \sap comes if we can balance the number of samples drawn in a way that negligibly impacts accuracy 
but still confers robustness against adversarial attacks.

\sap is similar to the \emph{dropout} technique due to \citet{srivastava2014dropout}. 
However, there is a crucial difference: 
\sap is more likely to sample activations 
that are high in absolute value, 
whereas dropout samples each activation with the same probability. 
Because of this difference, \sap, unlike dropout, 
can be applied post-hoc 
\emph{to pretrained models} 
without significantly decreasing 
the accuracy of the model. 
Experiments comparing \sap and dropout are included in section \ref{sec:experiments}.
Interestingly, dropout confers little advantage over the baseline.
We suspect that the reason for this is 
that the dropout training procedure 
encourages all possible dropout masks 
to result in similar mappings.

\subsection{Adversarial attack on \sap}
\label{stochasticAttack}
If the adversary knows that our defense policy
is to apply \sap,
it might try to calculate 
the best strategy against the \sap model.
Given the neural network $h$, input data $x$, 
corresponding true output $y$, 
a policy $\rho$ over the allowed perturbations, 
and a policy $\pi$ over the model parameters 
that come from \sap 
(this result holds true for any stochastic policy chosen over the model parameters), 
the adversary determines the optimal policy $\rho^*$
\begin{align*}
\rho^*=\arg\max_{\rho}\mathbb{E}_{p\sim\pi,r\sim\rho}[J(M_{p}(\theta),x+r,y)].
\end{align*}
Therefore, using the result from section \ref{sec:preliminaries}, the adversary determines the perturbation 
$\Delta x$ as follows;
\begin{align*}
\Delta x=\arg\max_{r}r^{\top}\mathbb{E}_{p\sim\pi}[\mathcal{J}(M_{p}(\theta),x,y)].
\end{align*}
To maximize the term, the adversary will set $r$ to be in the direction of sign of $\mathbb{E}_{p\sim\pi}[\mathcal{J}(M_{p}(\theta),x,y)]$.
Analytically computing $\mathbb{E}_{p\sim\pi}[\mathcal{J}(M_{p}(\theta),x,y)]$ is not feasible. 
However, the adversary can use Monte Carlo (MC) sampling 
to estimate the expectation 
as $\widetilde{\mathcal{J}}(M_{p}(\theta),x,y)$. 
Then, using FGSM,
$\Delta x=\lambda\cdot\text{sign}(\widetilde{\mathcal{J}}(M_{p}(\theta),x,y))$.

\section{Experiments}
\label{sec:experiments}
Our experiments to evaluate \sap
address two tasks:
image classification and reinforcement learning.
We apply the method to the \textit{ReLU} activation maps 
at each layer of the pretrained neural networks.
To create adversarial examples in our evaluation, we use FGSM, $\Delta x=\lambda\cdot\text{sign}(\mathcal{J}(M_{p}(\theta),x,y))$.
For stochastic models, the adversary estimates $\mathcal{J}(M_{p}(\theta),x,y)$ using MC sampling 
unless otherwise mentioned.
All perturbations are applied to the pixel values of images, which normally take values in the range $0$-$255$.
So the fraction of perturbation with respect to the data's dynamic range would be $\frac{\lambda}{256}$.
To ensure that all images are valid,
even following perturbation,
we clip the resulting pixel values 
so that they remain within the range $[0,255]$.
In all plots, we consider perturbations of the following magnitudes $\lambda=\{0,1,2,4,8,16,32,64\}$.\footnote{All the implementations were coded in MXNet framework \citep{chen2015mxnet} and sample code is available at \url{https://github.com/Guneet-Dhillon/Stochastic-Activation-Pruning}}

To evaluate models in the image classification domain,
we look at two aspects:
the model accuracy for varying values of $\lambda$,
and the calibration of the models \citep{guo2017calibration}.
Calibration of a model is the relation
between the confidence level of the model's output
and its accuracy.
A linear calibration is ideal,
as it suggests that the accuracy of the model
is proportional to the confidence level of its output.
To evaluate models in the reinforcement learning domain,
we look at the average score
that each model achieves on the games played, 
for varying values of $\lambda$.
The higher the score,
the better is the model's performance.
Because the units of reward are arbitrary,
we report results in terms of the the relative percent change in rewards.
In both cases, the output of stochastic models are computed as an average over multiple forward passes.

\subsection{Adversarial attacks in image classification}
\label{exp:cifar10}

The CIFAR-$10$ dataset \citep{krizhevsky2009learning} was used for the image classification domain.
We trained a ResNet-$20$ model \citep{he2016deep} using SGD, 
with minibatches of size $512$, 
momentum of $0.9$, 
weight decay of $0.0001$, 
and a learning rate of $0.5$ 
for the first $100$ epochs, 
then $0.05$ for the next $30$ epochs,
and then $0.005$ for the next $20$ epochs.
This achieved an accuracy of $89.8\%$ with cross-entropy loss and \textit{ReLU} non-linearity.
For all the figures in this section, 
we refer to this model as the \emph{dense} model.
\begin{figure*}[t]
    \centering
    \begin{subfigure}[t]{0.32\textwidth}
        \centering
        \includegraphics[width=1.0\linewidth]{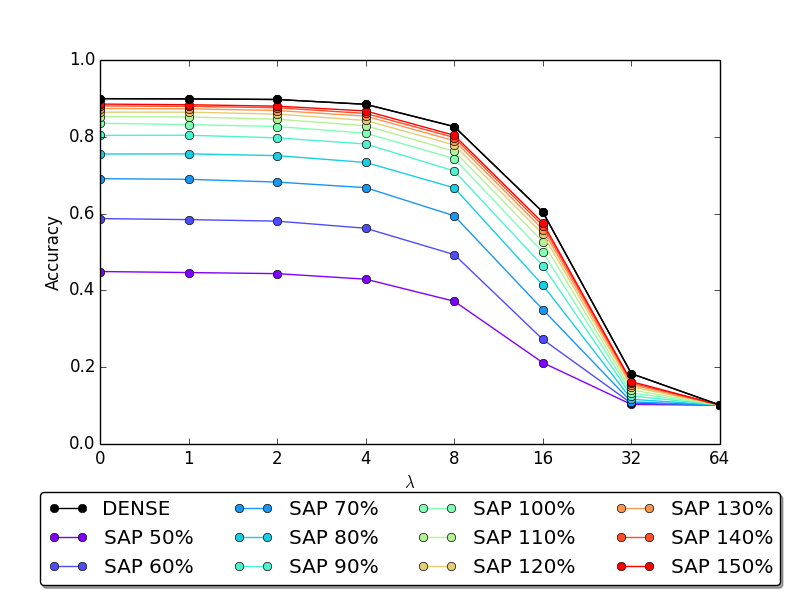}
        \vspace*{-0.5cm}
        \caption{}
        \label{fig:sap_random}
    \end{subfigure}
    \hfill
    \begin{subfigure}[t]{0.32\textwidth}
        \centering
        \includegraphics[width=1.0\linewidth]{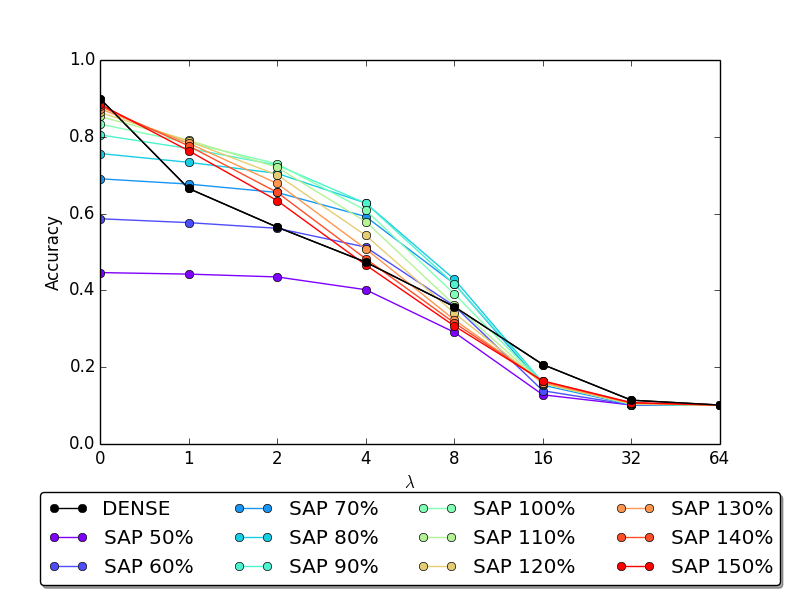}
        \vspace*{-0.5cm}
        \caption{}
        \label{fig:sap_sap}
    \end{subfigure}
    \hfill
    \begin{subfigure}[t]{0.32\textwidth}
        \centering
        \includegraphics[width=1.0\linewidth]{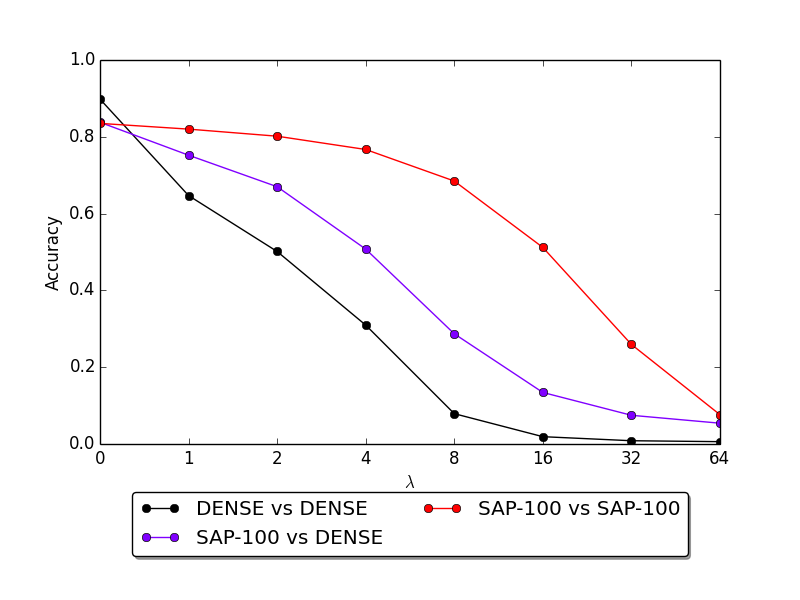}
        \vspace*{-0.5cm}
        \caption{}
        \label{fig:iterative}
    \end{subfigure}
    \caption{Accuracy plots of a variety of attacks against dense model and \sap models with different perturbation strengths, $\lambda$. For the \sap $\tau\%$ models, $\tau$ denotes the percentages of samples drawn from the multinomial distribution, at each layer. $(a)$ \sap models tested against random perturbation. $(b)$ \sap models tested against the FGSM attack, using MC sampling. $(c)$ \sap-$100$ tested against an iterative adversarial attack, using MC sampling (legend shows defender vs. adversary). It is worth restating that obtaining the iterative attack of \sap models is much more expensive and noisier than the iterative attack of dense models.}
\end{figure*}
The accuracy of the dense model degrades quickly with $\lambda$. 
For $\lambda=1$, the accuracy drops down to $66.3\%$, 
and for $\lambda=2$ it is $56.4\%$. 
These are small (hardly perceptible) perturbations in the input images, 
but the dense model's accuracy decreases
significantly.

\subsubsection{Stochastic activation pruning (\sap)}

We apply \sap to the dense model.
For each activation map $h^{i}\in\mathbb{R}^{a^{i}}$, 
we pick $k\%$ of $a^{i}$ activations to keep. Since activations are sampled with replacement, $k$ can be more than $100$. 
We will refer to $k$ as the percentage of samples drawn.
Fig. \ref{fig:sap_random} plots performance of \sap models against examples perturbed with random noise.
Perturbations of size $\lambda=64$ are readily perceptible and push all models under consideration
to near-random outputs, 
so we focus our attention on smaller values of $\lambda$.
Fig. \ref{fig:sap_sap} plots performance of these models against adversarial examples.
With many samples drawn, \sap converges to the dense model.
With few samples drawn, accuracy diminishes
for $\lambda=0$, but is higher for $\lambda\neq 0$.
The plot explains this balance well.
We achieve the best performance with $\sim100\%$ samples picked.
We will now only look at \sap $100\%$ (\sap-$100$). Against adversarial examples, with $\lambda=1$, $2$ and $4$, we observe a $12.2\%$, $16.3\%$ and $12.8\%$ absolute increase in accuracy respectively. However, for $\lambda=0$, we observe a $6.5\%$ absolute decrease in accuracy. For $\lambda=16$ again, there is a $5.2\%$ absolute decrease in accuracy.

\subsubsection{Dropout (DRO)}

\begin{figure*}[t]
    \centering
    \begin{subfigure}[t]{0.32\textwidth}
        \centering
        \includegraphics[width=1.0\linewidth]{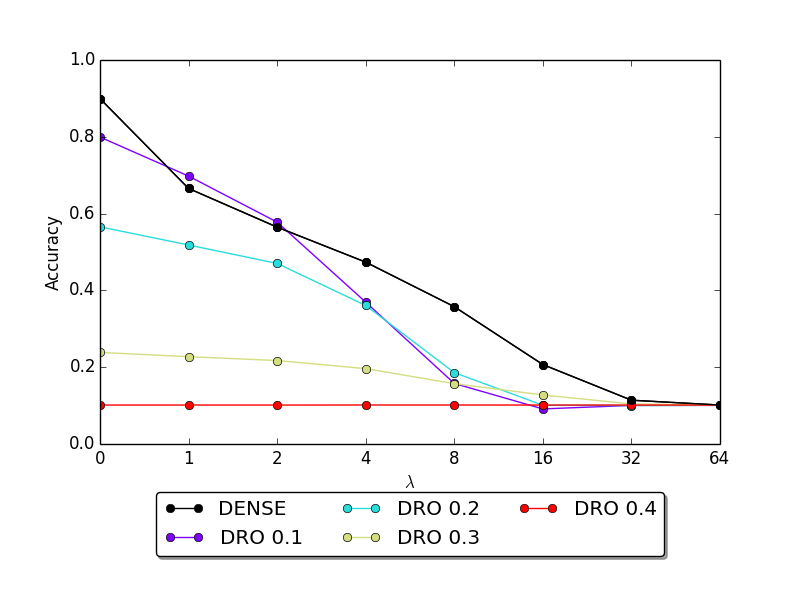}
        \vspace*{-0.7cm}
        \caption{}
    	\label{fig:DRO}
    \end{subfigure}
    \hfill
    \begin{subfigure}[t]{0.32\textwidth}
        \centering
        \includegraphics[width=1.0\linewidth]{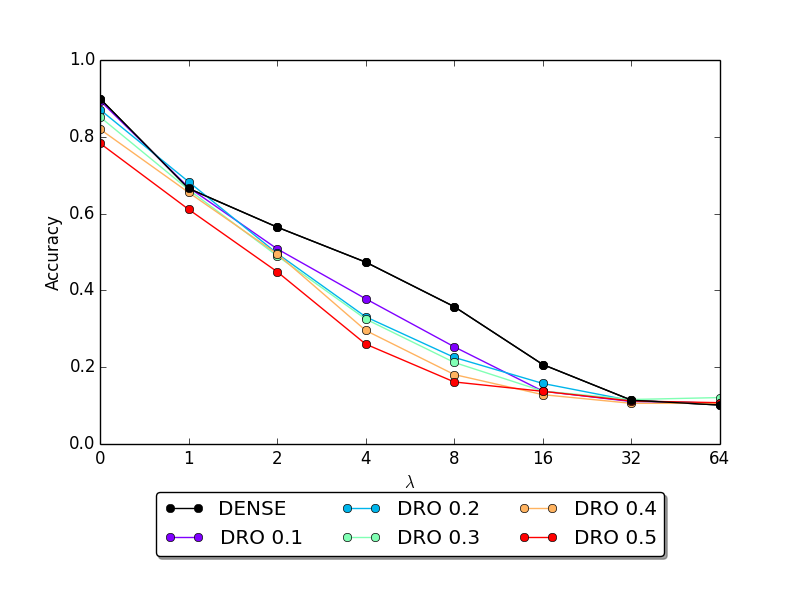}
        \vspace*{-0.7cm}
        \caption{}
    	\label{fig:DRO_STO}
    \end{subfigure}
    \hfill
    \begin{subfigure}[t]{0.32\textwidth}
        \centering
        \includegraphics[width=1.0\linewidth]{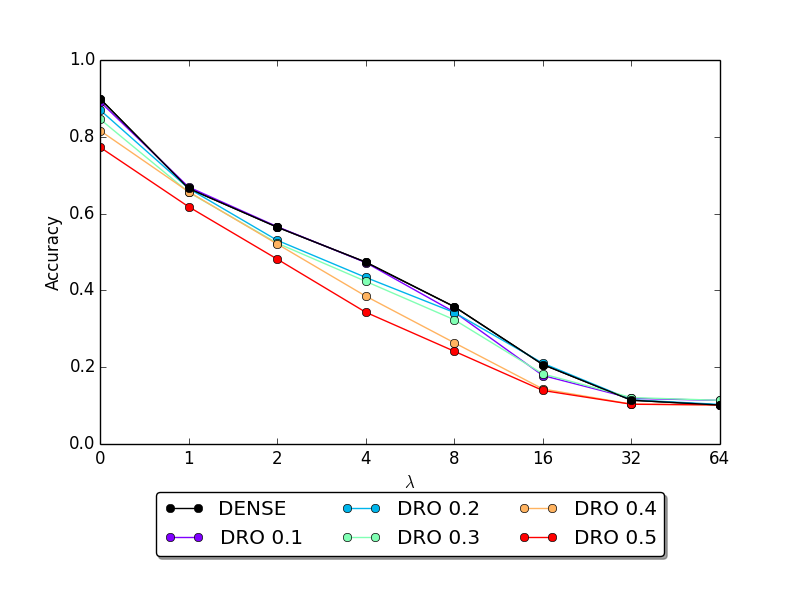}
        \vspace*{-0.7cm}
        \caption{}
    	\label{fig:DRO_NONSTO}
    \end{subfigure}
    \caption{Robustness of dropout models, with different rates of dropout (denoted in the legends), against adversarial attacks, using MC sampling, with a variety of perturbation strengths, $\lambda$: $(a)$ dropout is applied on the pre-trained models during the validations; $(b)$ the models are trained using dropout, and dropout is applied during the validations; $(c)$ the models are trained using dropout, but dropout is not applied during the validations.}
\end{figure*}

\emph{Dropout}, a technique due to \citet{srivastava2014dropout}, was also tested to compare with \sap.
Similar to the \sap setting, this method was added to the \textit{ReLU} activation maps of the dense model. We see that low dropout rate perform similar to the dense model for small $\lambda$ values, but its accuracy starts decreasing very quickly for higher $\lambda$ values (Fig. \ref{fig:DRO}).
We also trained ResNet-$20$ models, similar to the dense model, but with different dropout rates. This time, the models were trained for $250$ epochs, with an initial learning rate of $0.5$, reduced by a factor of $0.1$ after $100$, $150$, $190$ and $220$ epochs. These models were tested against adversarial examples with and without dropout during validation (Figs. \ref{fig:DRO_STO} and \ref{fig:DRO_NONSTO} respectively). The models do similar to the dense model, but do not provide additional robustness.

\subsubsection{Adversarial training (ADV)}

\emph{Adversarial training} \citep{goodfellow2014explaining} has emerged a standard method for defending against adversarial examples. It has been adopted by \citet{madry2017towards,tramer2017ensemble} to maintain high accuracy levels even for large $\lambda$ values.
We trained a ResNet-$20$ model, similar to the dense model, with an initial learning rate of $0.5$, which was halved every $10$ epochs, for a total of $100$ epochs.
It was trained on a dataset consisting of $80\%$ un-perturbed data and $20\%$ adversarially perturbed data, generated on the model from the previous epoch, with $\lambda=2$.
This achieved an accuracy of $75.0\%$ on the un-perturbed validation set.
Note that the model capacity was not changed. When tested against adversarial examples, the accuracy dropped to $72.9\%, 70.9\%$ and $67.5\%$ for $\lambda=1,2$ and $4$ respectively. We ran \sap-$100$ on the ADV model (referred to as ADV$+$\sap-$100$).
The accuracy in the no perturbation case was $74.1\%$.
For adversarial examples, both models act similar to each other for small values of $\lambda$. But for $\lambda=16$ and $32$, ADV$+$\sap-$100$ gets a higher accuracy than ADV by an absolute increase of $7.8\%$ and $7.9\%$ respectively.
%

\begin{figure}[t]
   \centering
   \includegraphics[width=0.5\linewidth]{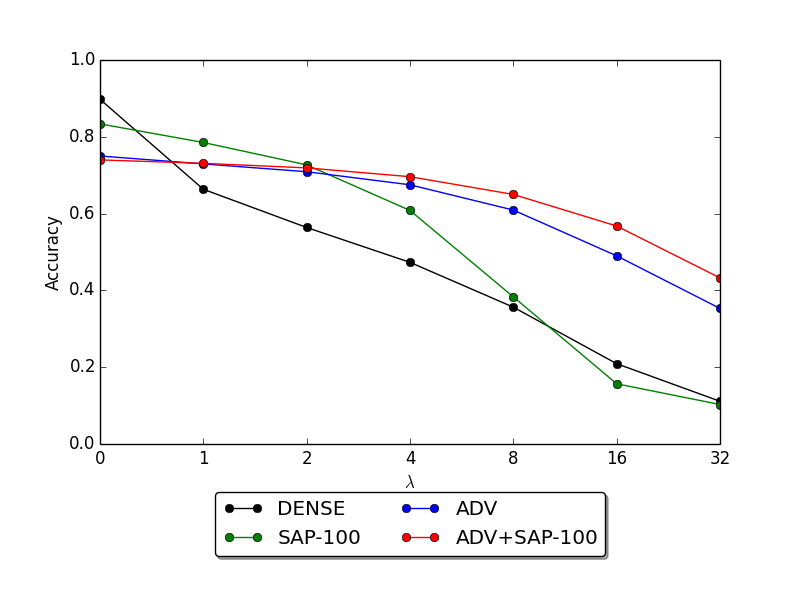}
   \caption{Accuracy plots of the dense, \sap-$100$, ADV and ADV$+$\sap-$100$ models, against adversarial attacks, using MC sampling, with a variety of perturbation strengths, $\lambda$.}
   \label{fig:accuracies}
\end{figure}

\begin{figure*}[t]
    \centering
    \begin{subfigure}[t]{0.32\textwidth}
        \centering
        \includegraphics[width=1.0\linewidth]{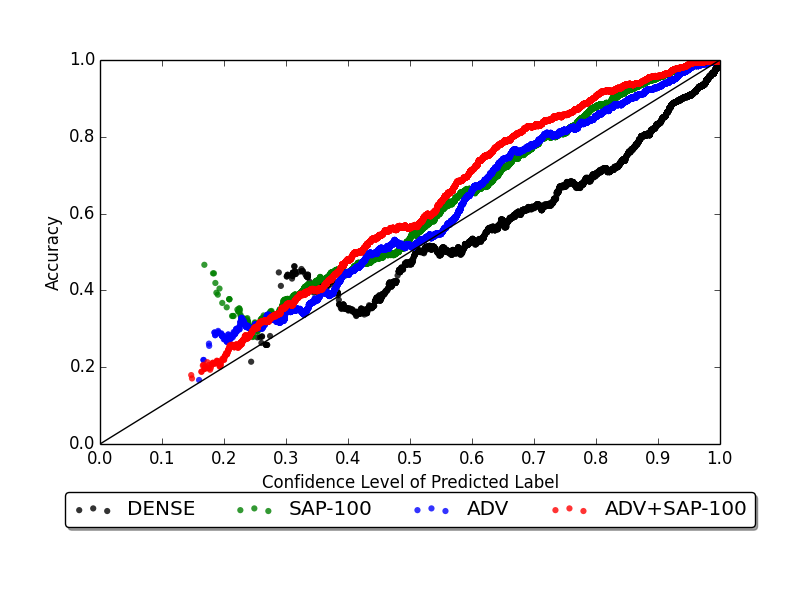}
        \vspace*{-0.9cm}
        \caption{$\lambda=0$.}
    \end{subfigure}
    \hfill
    \begin{subfigure}[t]{0.32\textwidth}
        \centering
        \includegraphics[width=1.0\linewidth]{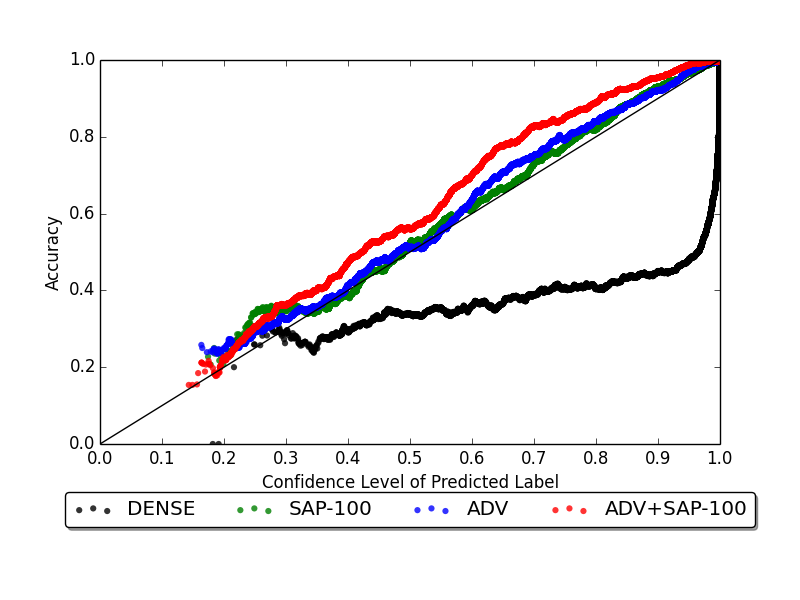}
        \vspace*{-0.9cm}
        \caption{$\lambda=1$.}
    \end{subfigure}
    \hfill
    \begin{subfigure}[t]{0.32\textwidth}
        \centering
        \includegraphics[width=1.0\linewidth]{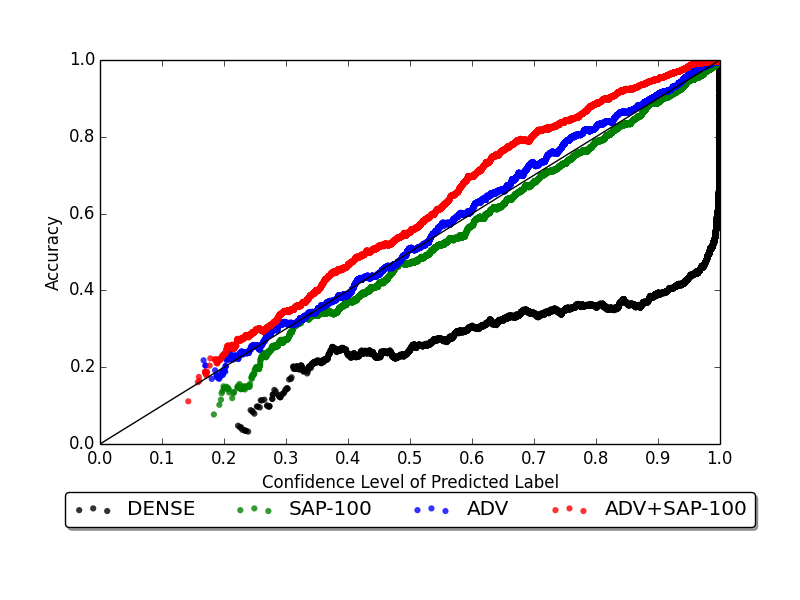}
        \vspace*{-0.9cm}
        \caption{$\lambda=2$.}
    \end{subfigure}
    \hfill
    \begin{subfigure}[t]{0.32\textwidth}
        \centering
        \includegraphics[width=1.0\linewidth]{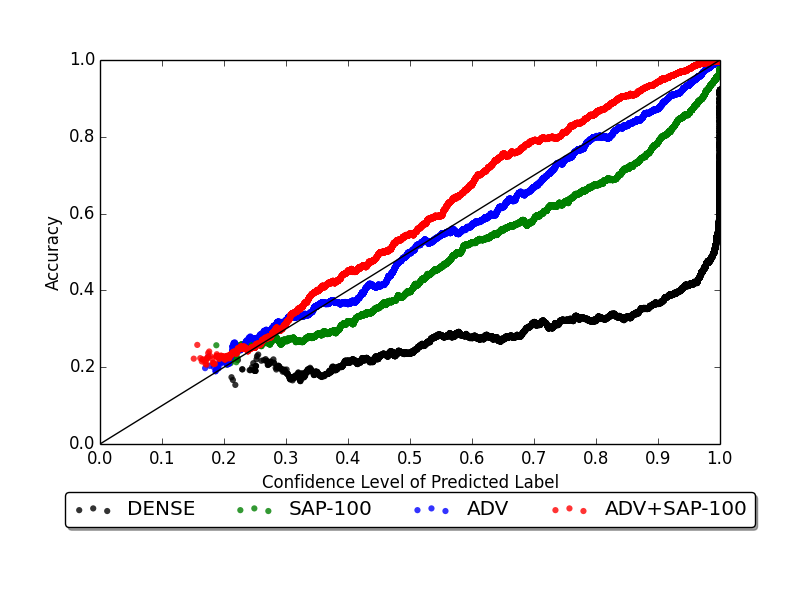}
        \vspace*{-0.9cm}
        \caption{$\lambda=4$.}
    \end{subfigure}
    \hfill
    \begin{subfigure}[t]{0.32\textwidth}
        \centering
        \includegraphics[width=1.0\linewidth]{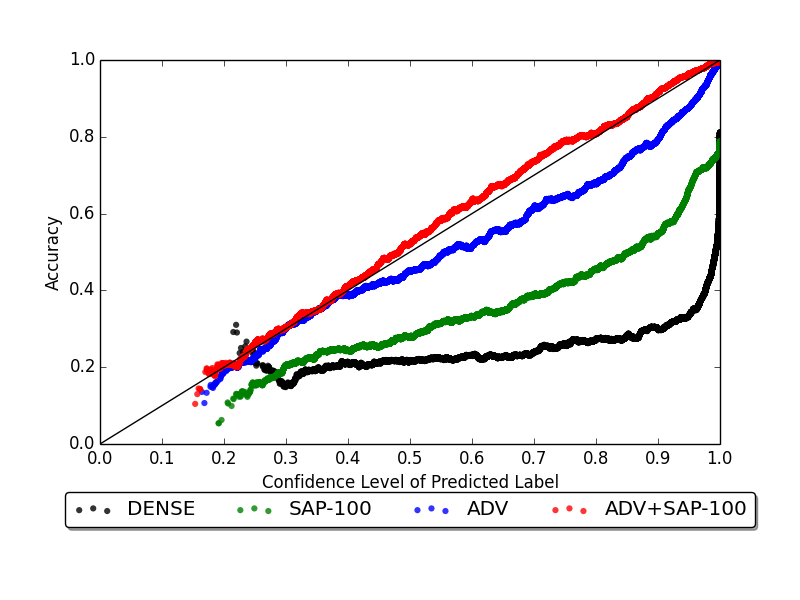}
        \vspace*{-0.9cm}
        \caption{$\lambda=8$.}
    \end{subfigure}
    \hfill
    \begin{subfigure}[t]{0.32\textwidth}
        \centering
        \includegraphics[width=1.0\linewidth]{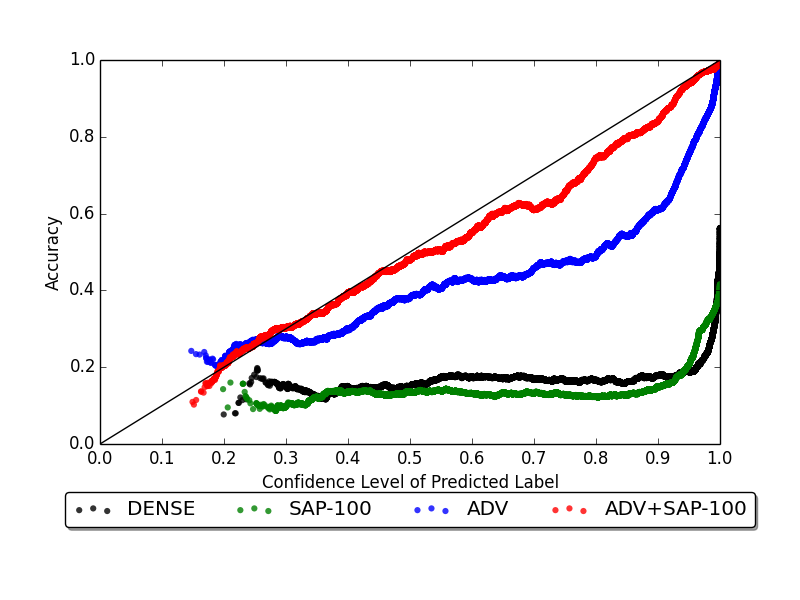}
        \vspace*{-0.9cm}
        \caption{$\lambda=16$.}
    \end{subfigure}
    \hfill
    \begin{subfigure}[t]{0.32\textwidth}
        \centering
        \includegraphics[width=1.0\linewidth]{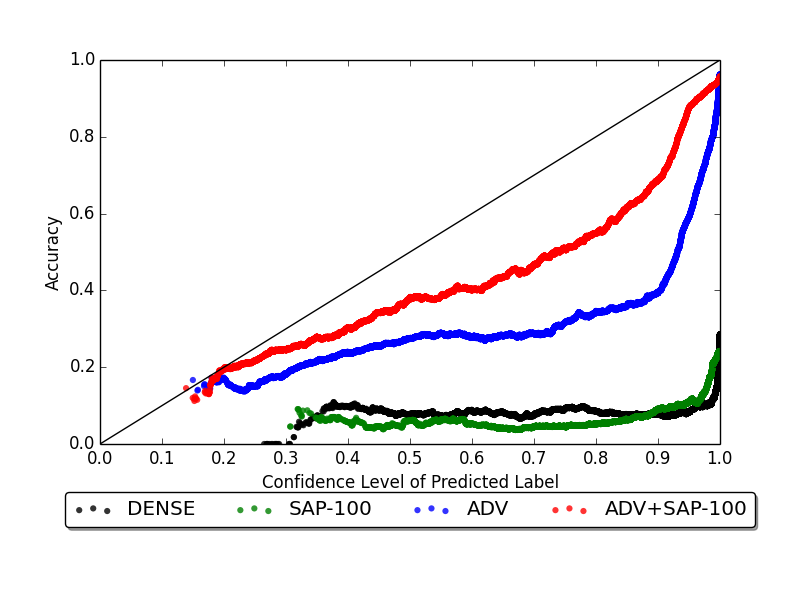}
        \vspace*{-0.9cm}
        \caption{$\lambda=32$.}
    \end{subfigure}
    \caption{Calibration plots of the dense, \sap-$100$, ADV and ADV$+$\sap-$100$ models, against adversarial attacks, using MC sampling, with a variety of different perturbation strengths, $\lambda$. These plots show the relation between the confidence level of the model's output and its accuracy.}
    \label{fig:calibration}
\end{figure*}

We compare the accuracy-$\lambda$ plot for dense, \sap-$100$, ADV and ADV$+$\sap-$100$ models. This is illustrated in Fig. \ref{fig:accuracies} For smaller values of $\lambda$, \sap-$100$ achieves high accuracy. As $\lambda$ gets larger, ADV$+$\sap-$100$ performs better than all the other models. We also compare the calibration plots for these models, in Fig. \ref{fig:calibration}. The dense model is not linear for any $\lambda\neq 0$.
The other models are well calibrated (close to linear), and behave similar to each other for $\lambda\leq 4$. For higher values of $\lambda$, we see that ADV$+$\sap-$100$ is the closest to a linearly calibrated model.

\subsection{Adversarial attacks in deep reinforcement learning (RL)}
\label{exp:rl}

Previously, \citep{behzadan2017vulnerability,huang2017adversarial,kos2017delving} 
have shown that the reinforcement learning agents 
can also be easily manipulated by adversarial examples. 
The RL agent learns the long term value $Q(a,s)$
of each state-action pair $(s,a)$
through interaction with an environment,
where given a state $s$, 
the optimal action is $\arg\max_a Q(a,s)$. 
A regression based algorithm, 
Deep Q-Network (DQN)\citep{mnih2015human} 
and an improved variant, 
Double DQN (DDQN) have been proposed 
for the popular Atari games \citep{bellemare2013arcade} as benchmarks. 
We deploy DDQN algorithm and train an RL agent in variety of different Atari game settings.

Similar to the image classification experiments, we tested \sap on a pretrained model (the model is described in the Appendix section \ref{app:rl}), by applying the method on the \textit{ReLU} activation maps. \sap-$100$ was used for these experiments.
Table \ref{table:rl} specifies the relative percentage increase in rewards of \sap-$100$ as compared to the original model.
For all the games, we observe a drop in performance for the no perturbation case. But for $\lambda\neq 0$, the relative increase in rewards is positive (except for $\lambda=1$ in the BattleZone game), and is very high in some cases ($3425.9\%$ for $\lambda=1$ for the Bowling game).

\begin{table}[t]
\caption {Relative percentage increase in rewards gained for \sap-$100$ compared to original model while playing different Atari games.}\label{table:rl}
\begin{center}
\scalebox{0.7}{
\begin{tabular}{c c c c c c c} 
    \toprule
    $\mathbf{\lambda}$ & \textbf{Assault} & \textbf{Asterix} & \textbf{BankHeist} & \textbf{BattleZone} & \textbf{BeamRider} & \textbf{Bowling} \\
    \midrule
    0 & -12.2\% & -33.4\% & -59.2\% & -65.8\% & -15.8\% & -4.5\% \\
    1 & 10.4\% & 13.3\% & 131.7\% & -22.0\% & 164.5\% & 3425.9\% \\
    2 & 9.8\% & 20.8\% & 204.8\% & 110.1\% & 92.3\% & \\
    4 & 12.4\% & 14.0\% & 1760.0\% & 202.6\% & & \\
    8 & 16.6\% & 7.4\% & 60.9\% & 134.8\% & & \\
    \bottomrule
\end{tabular}}
\end{center}
\end{table}

\subsection{Additional baselines}

In addition to experimenting with \sap, dropout, 
and adversarial training, 
we conducted extensive experiments with other methods 
for introducing stochasticity into a neural network. 
These techniques included $0$-mean Gaussian noise 
added to weights (RNW), $1$-mean multiplicative Gaussian noise for the weights (RSW), 
and corresponding additive (RNA) and multiplicative (RSA) noise added to the activations.
We describe each method in detail in
Appendix \ref{app:moreMethods}.
Each of these models
performs worse than the dense baseline
at most levels of perturbation 
and none matches the performance of \sap.
Precisely why SAP works while other methods
introducing stochasticity do not,
remains an open question that we continue to explore in future work.

\begin{figure*}[t]
    \centering
    \begin{subfigure}[t]{0.32\textwidth}
        \centering
        \includegraphics[width=1.0\linewidth]{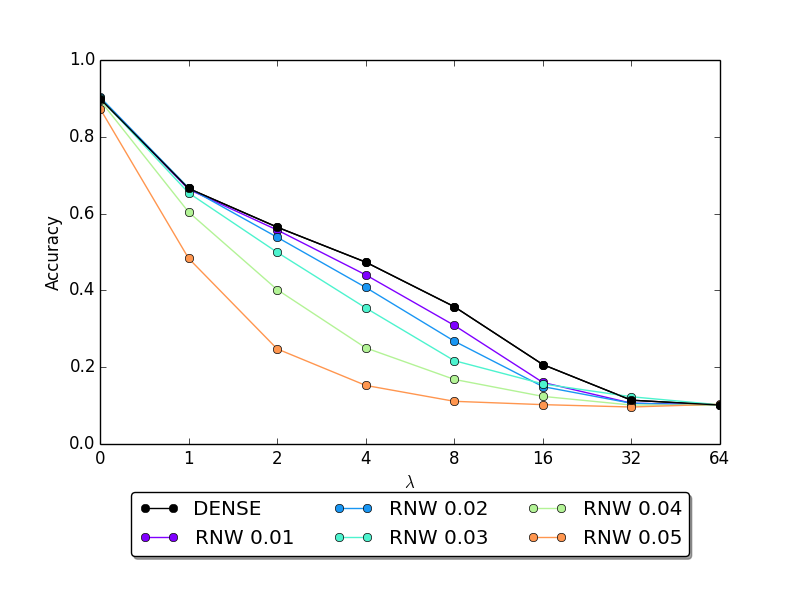}
        \vspace*{-0.5cm}
        \caption{Random noisy weights.}
        \label{fig:rnw}
    \end{subfigure}
    \hfill
    \begin{subfigure}[t]{0.32\textwidth}
        \centering
        \includegraphics[width=1.0\linewidth]{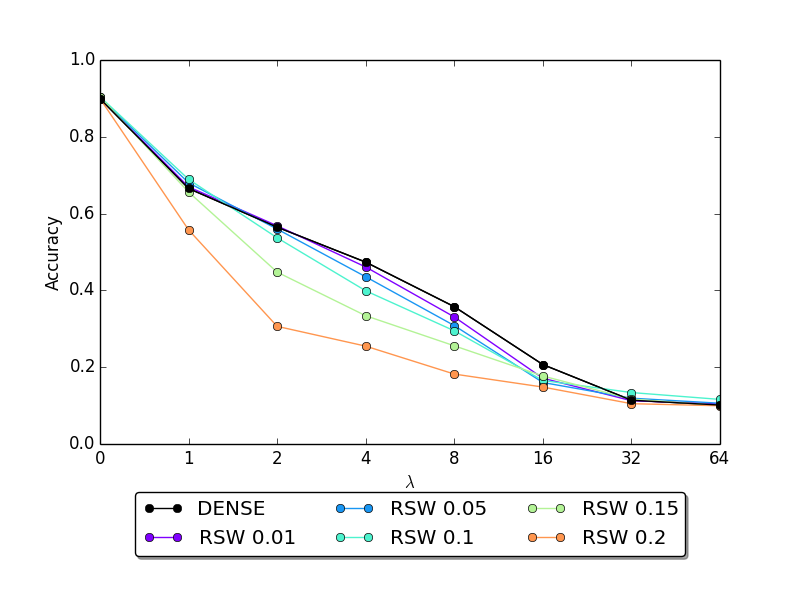}
        \vspace*{-0.5cm}
        \caption{Randomly scaled weights.}
        \label{fig:rsw}
    \end{subfigure}
    \hfill
    \begin{subfigure}[t]{0.32\textwidth}
        \centering
        \includegraphics[width=1.0\linewidth]{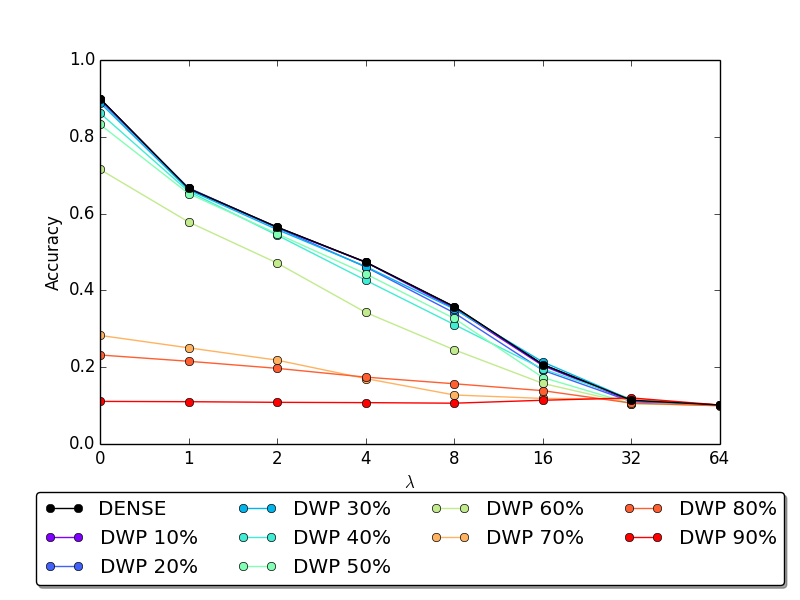}
        \vspace*{-0.5cm}
        \caption{Deterministic weight pruning.}
        \label{fig:dwp}
    \end{subfigure}
    \hfill
    \begin{subfigure}[t]{0.32\textwidth}
        \centering
        \includegraphics[width=1.0\linewidth]{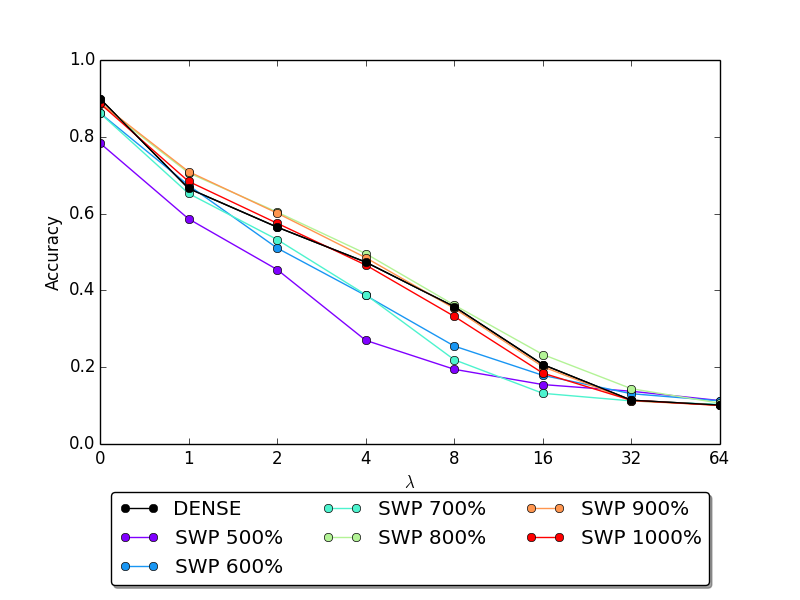}
        \vspace*{-0.5cm}
        \caption{Stochastic weight pruning.}
        \label{fig:swp}
    \end{subfigure}
    \hfill
    \begin{subfigure}[t]{0.32\textwidth}
        \centering
        \includegraphics[width=1.0\linewidth]{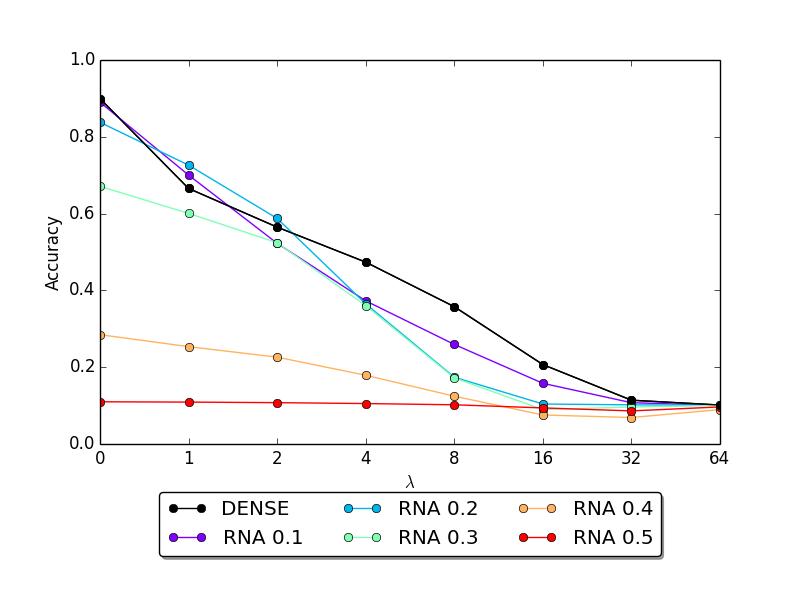}
        \vspace*{-0.5cm}
        \caption{Random noisy activations.}
        \label{fig:rna}
    \end{subfigure}
    \hfill
    \begin{subfigure}[t]{0.32\textwidth}
        \centering
        \includegraphics[width=1.0\linewidth]{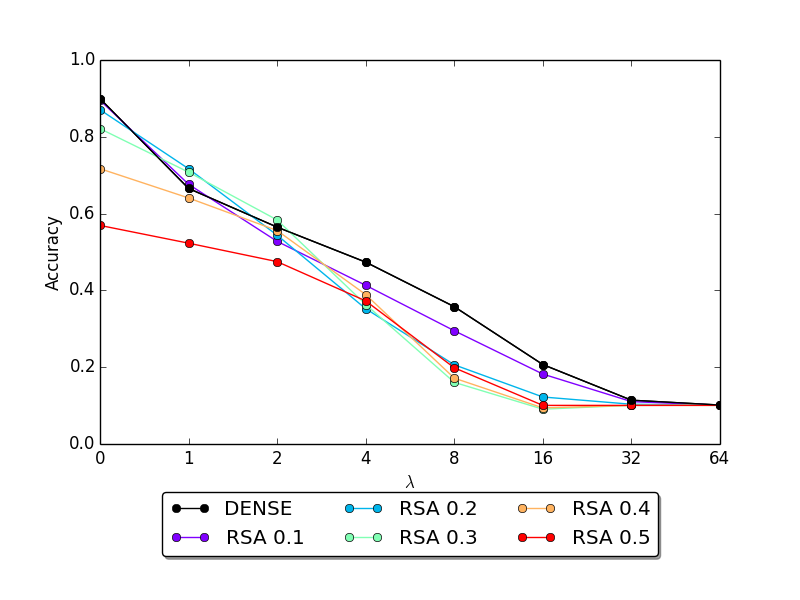}
        \vspace*{-0.5cm}
        \caption{Randomly scaled activations.}
        \label{fig:rsa}
    \end{subfigure}
    \caption{Robustness of 
different pruning and noisifying strategies against their respective adversarial attacks (MC sampling used to estimate gradients of stochastic models).}
\end{figure*}

\subsection{\sap attacks with varying numbers of MC samples}

In the previous experiments the \sap adversary used $100$ MC samples 
to estimate the gradient.
Additionally, we compared the performance of \sap-$100$ against various attacks, these include the standard attack calculated based on the  dense model and those generated on \sap-$100$ by estimating the gradient with various numbers of MC samples.
We see that if the adversary uses the dense model to generate adversarial examples, \sap-$100$ model's accuracy decreases.
Additionally, if the adversary uses the \sap-$100$ model to generate adversarial examples, greater numbers of MC samples lower the accuracy more.
Still, even with $1000$ MC samples, for low amounts of perturbation ($\lambda=1$ and $2$), \sap-$100$ retains higher accuracy than the dense model.

Computing a single backward pass of the \sap-$100$ model for $512$ examples takes $\sim 20$ seconds on $8$ GPUs.
Using $100$ and $1000$ MC samples
would take $\sim 0.6$ and $\sim 6$ hours respectively.

\subsection{Iterative adversarial attack}

A more sophisticated technique for producing adversarial  perturbations (than FGSM) is to apply multiple and smaller updates to the input in the direction of the local sign-gradients. 
This can be done by taking small steps of size $k\leq \lambda$ in the direction of the sign-gradient 
at the updated point and repeating the procedure
$\lceil\frac{\lambda}{k}\rceil$ times \citep{kurakin2016adversarial} as follows
\begin{align*}
x^{0}=x~~,\quad\quad
x^{t}=\textit{clip}_{x,\lambda}\left(x^{t-1}+k~\text{sign}(\mathcal{J}(\theta,x^{t-1},y))\right),\\
\end{align*}
where function $\textit{clip}_{x,\lambda}$ is a projection into a $L_{\infty}$-ball of radius $\lambda$ centered at $x$, and also into the hyper-cube of image space (each pixel is clipped to the range of $\left[0,255\right]$). 
The dense and \sap-$100$ models are tested against this adversarial attack, with $k=1.0$ (Fig. \ref{fig:iterative}).
The accuracies of the dense model at $\lambda=0$, $1$, $2$ and $4$ are $89.8\%$, $66.3\%$, $50.1\%$ and $31.0\%$ respectively.
The accuracies of the \sap-$100$ model against attacks  
computed on the same model (with $10$ MC samples taken at each step to estimate the gradient) 
are $83.3\%$, $82.0\%$, $80.2\%$ and $76.7\%$, 
for $\lambda=0, 1, 2,4$ respectively.
The \sap-$100$ model provides accuracies of $83.3\%$, $75.2\%$, $67.0\%$ and $50.8\%$, 
against attacks computed on the dense model, with the  perturbations $\lambda=0, 1, 2,4$ respectively. 
Iterative attacks on the \sap models 
are much more expensive to compute and noisier
than iterative attacks on dense models. 
This is why the adversarial attack computed on the dense model results in lower accuracies on the \sap-$100$ model than the adversarial attack computed on the \sap-$100$ model itself.

\begin{figure}[t]
   \centering
   \includegraphics[width=0.32\linewidth]{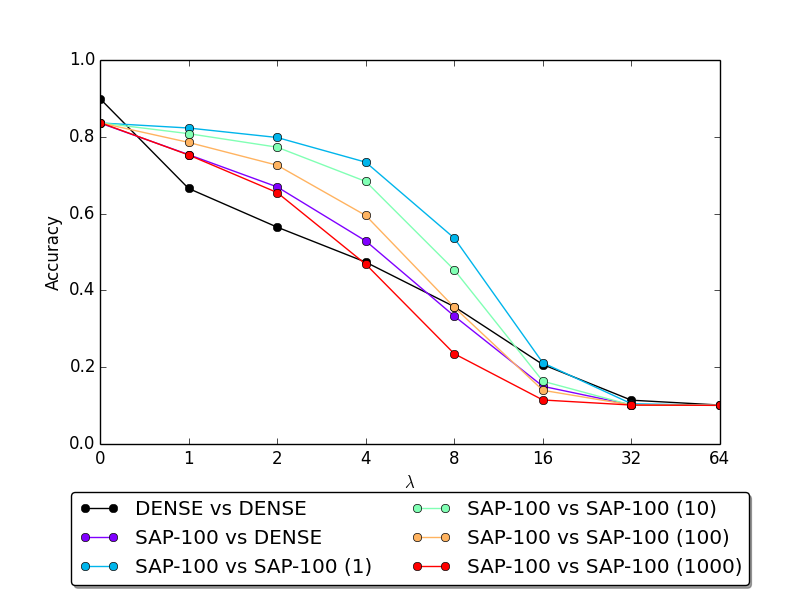}
   \caption{Accuracy plots of adversarial attacks, with different perturbation strengths, $\lambda$. The legend shows defender vs. adversary models (used for gradient computation), and the number of MC samples used to estimate the gradient.}
   \label{fig:mc_samples}
\end{figure}

\section{Related work}
\label{sec:related}
Robustness to adversarial attack has recently emerged as a serious topic in machine learning \citep{goodfellow2014explaining,kurakin2016adversarial,papernot2016effectiveness,tramer2017space, fawzi2018analysis}.
\citet{goodfellow2014explaining} introduced FGSM.
\citet{kurakin2016adversarial} proposed an iterative method where
FGSM
is used for smaller step sizes, which leads to a better approximation of the gradient.
\citet{papernot2017practical} observed that adversarial examples could be transferred to other models as well.
\citet{madry2017towards} introduce adding random noise to the image and then using the
FGSM method
to come up with adversarial examples.

Being robust against adversarial examples has primarily focused on training on the adversarial examples.
\citet{goodfellow2014explaining} use FGSM
to inject adversarial examples into their training dataset.
\citet{madry2017towards} use an iterative
FGSM approach to create adversarial examples to train on.
\citet{tramer2017ensemble} introduced an ensemble adversarial training method of training on the adversarial examples created on the model itself and an ensemble of other pre-trained models. These works have been successful, achieving only a small drop in accuracy form the clean and adversarially generated data.
\citet{nayebi2017biologically} proposes a method to produce a smooth input-output mapping by using saturating activation functions and causing the activations to become saturated.


\section{Conclusion}
\label{sec:discussion}
The \sap approach guards networks against adversarial examples 
without requiring any additional training.
We showed that in the adversarial setting, 
applying \sap to image classifiers improves both the accuracy and calibration.
Notably, combining \sap with adversarial training yields additive benefits.
Additional experiments show that \sap can also be effective against adversarial examples 
in reinforcement learning.

\bibliography{iclr2018_conference}

\begin{thebibliography}{22}
\providecommand{\natexlab}[1]{#1}
\providecommand{\url}[1]{\texttt{#1}}
\expandafter\ifx\csname urlstyle\endcsname\relax
  \providecommand{\doi}[1]{doi: #1}\else
  \providecommand{\doi}{doi: \begingroup \urlstyle{rm}\Url}\fi

\bibitem[Behzadan \& Munir(2017)Behzadan and Munir]{behzadan2017vulnerability}
Vahid Behzadan and Arslan Munir.
\newblock Vulnerability of deep reinforcement learning to policy induction
  attacks.
\newblock \emph{arXiv preprint arXiv:1701.04143}, 2017.

\bibitem[Bellemare et~al.(2013)Bellemare, Naddaf, Veness, and
  Bowling]{bellemare2013arcade}
Marc~G Bellemare, Yavar Naddaf, Joel Veness, and Michael Bowling.
\newblock The arcade learning environment: An evaluation platform for general
  agents.
\newblock \emph{J. Artif. Intell. Res.(JAIR)}, 47:\penalty0 253--279, 2013.

\bibitem[Chen et~al.(2015)Chen, Li, Li, Lin, Wang, Wang, Xiao, Xu, Zhang, and
  Zhang]{chen2015mxnet}
Tianqi Chen, Mu~Li, Yutian Li, Min Lin, Naiyan Wang, Minjie Wang, Tianjun Xiao,
  Bing Xu, Chiyuan Zhang, and Zheng Zhang.
\newblock Mxnet: A flexible and efficient machine learning library for
  heterogeneous distributed systems.
\newblock \emph{arXiv preprint arXiv:1512.01274}, 2015.

\bibitem[Fawzi et~al.(2018)Fawzi, Fawzi, and Frossard]{fawzi2018analysis}
Alhussein Fawzi, Omar Fawzi, and Pascal Frossard.
\newblock Analysis of classifiers’ robustness to adversarial perturbations.
\newblock \emph{Machine Learning}, 107\penalty0 (3):\penalty0 481--508, 2018.

\bibitem[Goodfellow et~al.(2014)Goodfellow, Shlens, and
  Szegedy]{goodfellow2014explaining}
Ian~J Goodfellow, Jonathon Shlens, and Christian Szegedy.
\newblock Explaining and harnessing adversarial examples.
\newblock \emph{arXiv preprint arXiv:1412.6572}, 2014.

\bibitem[Guo et~al.(2017)Guo, Pleiss, Sun, and Weinberger]{guo2017calibration}
Chuan Guo, Geoff Pleiss, Yu~Sun, and Kilian~Q Weinberger.
\newblock On calibration of modern neural networks.
\newblock \emph{arXiv preprint arXiv:1706.04599}, 2017.

\bibitem[Han et~al.(2015)Han, Mao, and Dally]{han2015deep}
Song Han, Huizi Mao, and William~J Dally.
\newblock Deep compression: Compressing deep neural networks with pruning,
  trained quantization and huffman coding.
\newblock \emph{arXiv preprint arXiv:1510.00149}, 2015.

\bibitem[He et~al.(2016)He, Zhang, Ren, and Sun]{he2016deep}
Kaiming He, Xiangyu Zhang, Shaoqing Ren, and Jian Sun.
\newblock Deep residual learning for image recognition.
\newblock In \emph{Proceedings of the IEEE conference on computer vision and
  pattern recognition}, pp.\  770--778, 2016.

\bibitem[Huang et~al.(2017)Huang, Papernot, Goodfellow, Duan, and
  Abbeel]{huang2017adversarial}
Sandy Huang, Nicolas Papernot, Ian Goodfellow, Yan Duan, and Pieter Abbeel.
\newblock Adversarial attacks on neural network policies.
\newblock \emph{arXiv preprint arXiv:1702.02284}, 2017.

\bibitem[Kos \& Song(2017)Kos and Song]{kos2017delving}
Jernej Kos and Dawn Song.
\newblock Delving into adversarial attacks on deep policies.
\newblock \emph{arXiv preprint arXiv:1705.06452}, 2017.

\bibitem[Krizhevsky \& Hinton(2009)Krizhevsky and
  Hinton]{krizhevsky2009learning}
Alex Krizhevsky and Geoffrey Hinton.
\newblock Learning multiple layers of features from tiny images.
\newblock 2009.

\bibitem[Kurakin et~al.(2016)Kurakin, Goodfellow, and
  Bengio]{kurakin2016adversarial}
Alexey Kurakin, Ian Goodfellow, and Samy Bengio.
\newblock Adversarial examples in the physical world.
\newblock \emph{arXiv preprint arXiv:1607.02533}, 2016.

\bibitem[Madry et~al.(2017)Madry, Makelov, Schmidt, Tsipras, and
  Vladu]{madry2017towards}
Aleksander Madry, Aleksandar Makelov, Ludwig Schmidt, Dimitris Tsipras, and
  Adrian Vladu.
\newblock Towards deep learning models resistant to adversarial attacks.
\newblock \emph{arXiv preprint arXiv:1706.06083}, 2017.

\bibitem[Mnih et~al.(2015)Mnih, Kavukcuoglu, Silver, Rusu, Veness, Bellemare,
  Graves, Riedmiller, Fidjeland, Ostrovski, et~al.]{mnih2015human}
Volodymyr Mnih, Koray Kavukcuoglu, David Silver, Andrei~A Rusu, Joel Veness,
  Marc~G Bellemare, Alex Graves, Martin Riedmiller, Andreas~K Fidjeland, Georg
  Ostrovski, et~al.
\newblock Human-level control through deep reinforcement learning.
\newblock \emph{Nature}, 518\penalty0 (7540):\penalty0 529--533, 2015.

\bibitem[Nayebi \& Ganguli(2017)Nayebi and Ganguli]{nayebi2017biologically}
Aran Nayebi and Surya Ganguli.
\newblock Biologically inspired protection of deep networks from adversarial
  attacks.
\newblock \emph{arXiv preprint arXiv:1703.09202}, 2017.

\bibitem[Osborne \& Rubinstein(1994)Osborne and Rubinstein]{osborne1994course}
Martin~J Osborne and Ariel Rubinstein.
\newblock \emph{A course in game theory}.
\newblock MIT press, 1994.

\bibitem[Papernot \& McDaniel(2016)Papernot and
  McDaniel]{papernot2016effectiveness}
Nicolas Papernot and Patrick McDaniel.
\newblock On the effectiveness of defensive distillation.
\newblock \emph{arXiv preprint arXiv:1607.05113}, 2016.

\bibitem[Papernot et~al.(2017)Papernot, McDaniel, Goodfellow, Jha, Celik, and
  Swami]{papernot2017practical}
Nicolas Papernot, Patrick McDaniel, Ian Goodfellow, Somesh Jha, Z~Berkay Celik,
  and Ananthram Swami.
\newblock Practical black-box attacks against machine learning.
\newblock In \emph{Proceedings of the 2017 ACM on Asia Conference on Computer
  and Communications Security}, pp.\  506--519. ACM, 2017.

\bibitem[Srivastava et~al.(2014)Srivastava, Hinton, Krizhevsky, Sutskever, and
  Salakhutdinov]{srivastava2014dropout}
Nitish Srivastava, Geoffrey~E Hinton, Alex Krizhevsky, Ilya Sutskever, and
  Ruslan Salakhutdinov.
\newblock Dropout: a simple way to prevent neural networks from overfitting.
\newblock \emph{Journal of machine learning research}, 15\penalty0
  (1):\penalty0 1929--1958, 2014.

\bibitem[Szegedy et~al.(2013)Szegedy, Zaremba, Sutskever, Bruna, Erhan,
  Goodfellow, and Fergus]{szegedy2013intriguing}
Christian Szegedy, Wojciech Zaremba, Ilya Sutskever, Joan Bruna, Dumitru Erhan,
  Ian Goodfellow, and Rob Fergus.
\newblock Intriguing properties of neural networks.
\newblock \emph{arXiv preprint arXiv:1312.6199}, 2013.

\bibitem[Tram{\`e}r et~al.(2017{\natexlab{a}})Tram{\`e}r, Kurakin, Papernot,
  Boneh, and McDaniel]{tramer2017ensemble}
Florian Tram{\`e}r, Alexey Kurakin, Nicolas Papernot, Dan Boneh, and Patrick
  McDaniel.
\newblock Ensemble adversarial training: Attacks and defenses.
\newblock \emph{arXiv preprint arXiv:1705.07204}, 2017{\natexlab{a}}.

\bibitem[Tram{\`e}r et~al.(2017{\natexlab{b}})Tram{\`e}r, Papernot, Goodfellow,
  Boneh, and McDaniel]{tramer2017space}
Florian Tram{\`e}r, Nicolas Papernot, Ian Goodfellow, Dan Boneh, and Patrick
  McDaniel.
\newblock The space of transferable adversarial examples.
\newblock \emph{arXiv preprint arXiv:1704.03453}, 2017{\natexlab{b}}.

\end{thebibliography}
\bibliographystyle{iclr2018_conference}

\clearpage
\appendix
\section{Reinforcement learning model architecture}
\label{app:rl}

For the experiments in section \ref{exp:rl}, we trained the network with RMSProp, minibatches of size $32$, a learning rate of $0.00025$, and a momentum of $0.95$ and as in \citep{mnih2015human} where the discount factor is $\gamma = 0.99$, the number of steps between target updates to $10000$ steps. We updated the network every $4$ steps by randomly sampling a minibatch of size $32$ samples from the replay buffer and trained the agents for a total of $100M$ steps per game. The experience replay contains the $1M$ most recent transitions. For training we used an $\varepsilon$-greedy policy with $\varepsilon$ annealed linearly from $1$ to $0.1$ over the first $1M$ time steps and fixed at $0.1$ thereafter.

The input to the network is $4\times 84\times 84$ tensor with a rescaled, gray-scale version of the last four observations. The first convolution layer has 32 filters of size $8$ with a stride of $4$. The second convolution layer has $64$ filters of size $4$ with stride $2$. The last convolution layer has $64$ filters of size $3$ followed by two fully connected layers with size $512$ and the final fully connected layer Q-value of each action where \textit{ReLU} rectifier is deployed for the nonlinearity at each layer.

\section{Other methods}
\label{app:moreMethods}

We tried a variety of different methods that could be added to pretrained models and tested their performance against adversarial examples.
The following is a continuation of section \ref{exp:cifar10}, where we use the dense model again on the CIFAR-$10$ dataset.

\subsection{Random noisy weights (RNW)}

One simple way of introducing stochasticity to the activations is by adding random Gaussian noise to each weight, with mean $0$ and constant standard deviation, $s$. So each weight tensor $W^{i}$ now changes to $M(W^{i})$, where the $j$'th entry is given by
\begin{align*}
M(W^{i})_{j}=(W^{i})_{j}+\eta,&&\eta\sim \mathcal{N}(0, s^{2}).
\end{align*}

These models behave very similar to the dense model (Fig. \ref{fig:rnw}, the legend indicates the value of $s$). 
While we test several different values of $s$, we do not observe any significant improvements regarding robustness against adversarial examples. 
As $s$ increased, the accuracy for non-zero $\lambda$ decreased.

\subsection{Randomly scaled weights (RSW)}

Instead of using additive noise, we also try multiplicative noise. The scale factor can be picked from a Gaussian distribution, with mean $1$ and constant standard deviation $s$. So each weight tensor $W^{i}$ now changes to $M(W^{i})$, where the $j$'th entry is given by
\begin{align*}
M(W^{i})_{j}=\eta\cdot(W^{i})_{j},&&\eta\sim \mathcal{N}(1, s^{2}).
\end{align*}

These models perform similar to the dense model, but again, no robustness is offered against adversarial examples.
They follow a similar trend as the RNW models (Figure \ref{fig:rsw}, the legend indicates the value of $s$).

\subsection{Deterministic weight pruning (DWP)}

Following from the motivation of preventing perturbations to propagate forward in the network, we tested deterministic weight pruning, where the top $k\%$ entries of a weight matrix were kept, while the rest were pruned to $0$, according to their absolute values. This method was prompted by the success achieved by this pruning method, introduced by \cite{han2015deep}, where they also fine-tuned the model.

For low levels of pruning, these models do very similar to the dense model, even against adversarial examples (Fig. \ref{fig:dwp}, the legend indicates the value of $k$). The adversary can compute the gradient of the sparse model, and the perturbations propagate forward through the surviving weights. For higher levels of sparsity, the accuracy in the no-perturbation case drops down quickly.

\subsection{Stochastic weight pruning (SWP)}

Observing the failure of deterministic weight pruning, we tested a mix of stochasticity and pruning, the stochastic weight pruning method. Very similar to the idea of \sap, we consider all the entries of a weight tensor to be a multinomial distribution, and we sample from it with replacement. For a weight tensor $W^{i}\in\mathbb{R}^{a^{i}}$, we sample from it $r^{i}$ times with replacement. The probability of sampling $(W^{i})_{j}$ is given by
\begin{align*}
p^{i}_{j}=\frac{|(W^{i})_{j}|}{\sum_{k=1}^{a^{i}}|(W^{i}))_{k}|}.
\end{align*}
The new weight entry, $M(W^{i})_{j}$, is given by
\begin{align*}
M(W^{i})_{j}=W^{i}_{j}\cdot\frac{\mathbb{I}(W^{i}_{j})}{1-(1-p_{i})^{r^{i}}},
\end{align*}
where $\mathbb{I}(W^{i}_{j})$ is the indicator function that returns $1$ if $W^{i}_{j}$ was sampled at least once, and $0$ otherwise.

For these experiments, for each weight matrix $W^{i}\in\mathbb{R}^{m^{i}}$, the number of samples picked were $k\%$ of $m^{i}$. Since samples were picked with replacement, $k$ could be more than $100$. We will refer to $k$ as the percentage of samples drawn.

These models behave very similar to the dense model. 
We tried drawing range of percentages of samples, but no evident robustness could be seen against adversarial examples (Figure \ref{fig:swp}, the legend indicates the value of $k$).
For a small $s$, it is very similar to the dense model. As $s$ increases, the these models do marginally better for low non-zero $\lambda$ values, and then drops again (similar to the \sap case).

\subsection{Random noisy activations (RNA)}

Next we change our attention to the activation maps in the dense model. One simple way of introducing stochasticity to the activations is by adding random Gaussian noise to each activation entry, with mean $0$ and constant standard deviation, $s$. So each activation map $h^{i}$ now changes to $M(h^{i})$, where the $j$'th entry is given by
\begin{align*}
M(h^{i})_{j}=(h^{i})_{j}+\eta,&&\eta\sim \mathcal{N}(0, s^{2}).
\end{align*}

These models too do not offer any robustness against adversarial examples. Their accuracy drops quickly with $\lambda$ and $s$ (Fig. \ref{fig:rna}, the legend indicates the value of $s$).

\subsection{Randomly scaled activations (RSA)}

Instead of having additive noise, we can also make the model stochastic by scaling the activations. The scale factor can be picked from a Gaussian distribution, with mean $1$ and constant standard deviation $s$. So each activation map $h^{i}$ now changes to $M(h^{i})$, where the $j$'th entry is given by
\begin{align*}
M(h^{i})_{j}=\eta(h^{i})_{j},&&\eta\sim \mathcal{N}(1, s^{2}).
\end{align*}

These models perform similar to the dense model, exhibiting no additional robustness against adversarial examples (Figure \ref{fig:rsa}, the legend indicates the value of $s$).

\end{document}